\documentclass[conference]{IEEEtran}
\ifCLASSINFOpdf
  \usepackage[pdftex]{graphicx}
  \graphicspath{ {pictures/}{../pdf/}{../jpeg/} }
  \DeclareGraphicsExtensions{.pdf,.jpeg,.png,.PNG}
\else
\fi

\usepackage{flushend}
\usepackage{url}
\usepackage[table]{xcolor}
\usepackage{tabularx,booktabs,textcomp}
\newcolumntype{C}{>{\centering\arraybackslash}X} % centered version of "X" type
\newcolumntype{L}{>{\raggedright\arraybackslash}X} % LEFT version... "\...right" works?
\usepackage[noadjust]{cite}
\usepackage{hhline}

\usepackage{soul}
\usepackage{xcolor}
\usepackage{subcaption}
\usepackage{multirow}
\usepackage{mwe}
\usepackage{subcaption}
\usepackage{graphicx}
\usepackage[vlined,ruled,linesnumbered]{algorithm2e}
\usepackage{wasysym}
\usepackage{lipsum} % USE: "\lipsum[1]" to generate 1 paragraph of random text 
\usepackage{array}
\usepackage{multirow}
\usepackage{listings}
\colorlet{punct}{red!60!black}
\definecolor{background}{HTML}{EEEEEE}
\definecolor{delim}{RGB}{20,105,176}
\colorlet{numb}{magenta!60!black}
\lstdefinelanguage{json}{
    basicstyle=\normalfont\ttfamily,
    numbers=left,
    numberstyle=\scriptsize,
    stepnumber=1,
    numbersep=8pt,
    showstringspaces=false,
    breaklines=true,
    frame=lines,
    backgroundcolor=\color{background},
    literate=
     *{0}{{{\color{numb}0}}}{1}
      {1}{{{\color{numb}1}}}{1}
      {2}{{{\color{numb}2}}}{1}
      {3}{{{\color{numb}3}}}{1}
      {4}{{{\color{numb}4}}}{1}
      {5}{{{\color{numb}5}}}{1}
      {6}{{{\color{numb}6}}}{1}
      {7}{{{\color{numb}7}}}{1}
      {8}{{{\color{numb}8}}}{1}
      {9}{{{\color{numb}9}}}{1}
      {:}{{{\color{punct}{:}}}}{1}
      {,}{{{\color{punct}{,}}}}{1}
      {\{}{{{\color{delim}{\{}}}}{1}
      {\}}{{{\color{delim}{\}}}}}{1}
      {[}{{{\color{delim}{[}}}}{1}
      {]}{{{\color{delim}{]}}}}{1},
}
\definecolor{codegreen}{rgb}{0,0.6,0}
\definecolor{codegray}{rgb}{0.5,0.5,0.5}
\definecolor{codepurple}{rgb}{0.58,0,0.82}
\definecolor{backcolour}{rgb}{0.95,0.95,0.92}
 
\newcommand\MyBox[2]{
  \fbox{\lower0.75cm
    \vbox to 1.7cm{\vfil
      \hbox to 1.7cm{\hfil\parbox{1.4cm}{#1\\#2}\hfil}
      \vfil}%
  }%
}
\begin{document}

\title{Task-Specific Activation Functions for Neuroevolution using Grammatical Evolution}

% AUTHOR
\author{\IEEEauthorblockN{Benjamin David Winter\IEEEauthorrefmark{1},
William John Teahan\IEEEauthorrefmark{2}}
\IEEEauthorblockA{School Of Computer Science and Electronic Engineering\\
Bangor University,
Wales\\
Email: \IEEEauthorrefmark{1}eeu60d@bangor.ac.uk,
\IEEEauthorrefmark{2}w.j.teahan@bangor.ac.uk}}
\maketitle

%\maketitle

\begin{abstract}
Activation functions play a critical role in the performance and behaviour of neural networks, significantly impacting their ability to learn and generalise. Traditional activation functions, such as ReLU, sigmoid, and tanh, have been widely used with considerable success. However, these functions may not always provide optimal performance for all tasks and datasets. In this paper, we introduce Neuvo GEAF -- an innovative approach leveraging grammatical evolution (GE) to automatically evolve novel activation functions tailored to specific neural network architectures and datasets. Experiments conducted on well-known binary classification datasets show statistically significant improvements in F1-score (between 2.4\% and 9.4\%) over ReLU using identical network architectures. Notably, these performance gains were achieved without increasing the network's parameter count, supporting the trend toward more efficient neural networks that can operate effectively on resource-constrained edge devices. This paper's findings suggest that evolved activation functions can provide significant performance improvements for binary classification networks.

\end{abstract}


\section{Introduction}
Choosing the right activation function for a neural network for any particular task or dataset can be a tedious process and may require substantial research as there is no rule of thumb for selecting an activation function because the choice of activation functions is dependent on the context/task \cite{sharma}.
The value from an activation function that is passed on to the loss function is extremely important. In a standard neural network, the activation function passes its value on to a loss function. A problem many researchers face is not knowing which activation function to use for each task. There are ten activation functions that are most prominent: \textit{Binary step}, \textit{Linear}, \textit{Sigmoid}, \textit{Tanh}, \textit{ReLU}, \textit{Leaky ReLU}, \textit{ELU}, \textit{Swish}, \textit{Softmax} and \textit{Softplus}. They all have their own issues and trade-offs, but their efficiency is dependent on the data/task they are performing.

However, a poorly chosen activation function can have detrimental effects on the network's results. The vanishing gradient problem is a phenomenon that occurs when inputs that are close to 1 automatically snap to 1 whilst inputs that are close to 0 automatically snap to 0 or -1 respectively \cite{brownlee}. This results in a lack of gradient for results, hence the name vanishing gradient'. Its opposite, the exploding gradient' is the result of the error gradients accumulating during an update and resulting in typically exponentially large error gradients each epoch. This in turn results in an unstable network and can even change the network's weights to NaN values. Furthermore, some of the activation functions previously mentioned are momentum based, resulting in a disparity in convergence time over a non-momentum based activation function. This in itself has a large effect on the time it takes to converge to a minima, as well as determining if the value will get stuck on a saddle point.
Various papers \cite{bingham, sharma} have shown that even if a catastrophic error is avoided in the selection of a neural network activation function, the neural network's accuracy can be vastly improved by the design of custom activation functions. These are activation functions that are not classed as typical and may only improve results for one specific dataset.
As the field of machine learning increasingly pushes toward deployment on edge devices and energy-efficient computing, optimizing smaller neural networks becomes critically important. While large language models (LLMs) excel at general tasks, they are impractical for many real-world applications requiring rapid, energy-efficient classification on resource-constrained hardware. The ability to achieve higher accuracy with compact networks through optimized activation functions represents a significant advancement for applications like medical diagnostics, sensor networks, and IoT devices, where both accuracy and efficiency are paramount concerns.

This paper outlines a novel method of creating custom activation functions through Grammatical Neuroevolution on a task-specific basis -- Neuvo GEAF, in which we create the search space using a Backas-Naur form grammar and recurse through the grammar to map evolvable genotypes to phenotypes which represent a neural network's activation functions. Our approach delivers substantial performance improvements without increasing model complexity or computational requirements during inference, making it particularly valuable for deployment scenarios where energy consumption and computational resources are limited.
The paper is organised as follows. Section two describes the background of Grammatical Evolution and its benefits over its competitor algorithms, whilst also discussing some of the more prominent Neuroevolution algorithms and models, specifically ones that feature Grammatical Neuroevolution or the creation of hand crafted activation functions. Section three summarises the grammar that is used for this project, and more specifically its usefulness in creating both emergent activation functions and well known activation functions. The methodology of the creation of this model regarding implementation is defined in section four of this paper. Section five regards the experimental setup, which datasets are used, the baseline network parameters used for fair testing as well as providing the hardware specifications used to make, train, validate and test the model. Section six discusses the results, the benefits and drawbacks of this approach and a general discussion surrounding the future of Grammatical Neuroevolution. Lastly, section seven concludes the paper and describes avenues that Grammatical Neuroevolution should focus on in the future.

\section{Background and Related Work}
\subsection{Grammatical Evolution}
Grammatical evolution (GE) is an evolutionary algorithm that can evolve complete programs in an arbitrary language using a variable-length string that represents the genome. The genome determines which production rules in a Backus-Naur form grammar definition are used in a genotype-to-phenotype mapping process to a program \cite{ONeil2003}.

GE has several benefits over Genetic Programming (GP), the first of which being an efficiency gain due to the ability to constrict the search space and thus avoid bloat. Bloat is typically defined as an increase in the size of the program without a corresponding improvement in fitness. Whilst many researchers have created models to combat bloat in GP, it still exists \cite{trujillo, whigham, poli, bleuler}.

A major benefit of GE is that it works in a way that is more similar to how the role of DNA in natural evolution works than other EA's, and therefore can exploit the natural divide between genotype and phenotype. Humans look and behave relatively similar to each other in comparison to other organisms. However, according to a DNA-sequencing analysis of the Y chromosome, it has been found that every time human DNA is passed from one generation to the next, it accumulates 100–200 new mutations \cite{dolgin}. Most of these mutations are known as silent mutations, which have no observable effect on the organism’s phenotype. GE utilises a concept called degeneracy which can produce these silent mutations within a genotype which have no effect on the phenotype \cite{oneill2}. 

A further feature of GE is deliberate bias, which allows the developer to create the grammar in such a way that certain features are more likely to be chosen. The grammar used in this project (seen in Table~\ref{fig:grammar}) was unbiased as there was an equal amount of chance for each feature. However, one could add deliberate bias by making one feature more likely to be chosen over its counterparts.  
\label{sec:ge}

Nicolau et al. \cite{nicolau} stated that to design a good grammar, deep knowledge of both GE and the application domain are required. However, they also allude to some general principles one should follow when designing an effective grammar. Firstly, the reduction of the number of non-terminal symbols within the grammar is particularly useful when using a linear genome representation. Secondly, they found that symbol biases can both degrade or improve performance between problems, concluding that symbol biases are problem dependant and should be carefully considered before a grammar is designed. 

\subsection{Grammatical Neuroevolution}
Grammar based neuroevolution or Grammatical Neuroevolution is usually defined as the modification of a neural network's topology, weights or parameters via the use of grammatical evolution. DSGE is a model developed by Assun{\c{c}}{\~{a}}o  et al. \cite{assuncao} and was one of the first models that used Grammatical Evolution for Neuroevolution to create dynamic rules that describe the placement and architecture of neurons in the network. DSGE was developed to solve the problem with Grammar-based Genetic Programming (GGP), which is that GGP is limited to the evolution of networks with only one hidden-layer. DSGE allows multiple hidden layers to be added through a recursive mapping process. DSGE also allows parameters such as the number of neurons per hidden layer and the weights of each connection to be placed into genotypes. 

Wong et al. \cite{wong} also created a Grammatical Neuroevolution solution. Their implementation was used to classify ventricular tachycardia signals. The problem they found was that traditional deep neural networks had to be developed by human engineers and finding the optimal amount of hidden layers was sometimes challenging. They developed ADAG-DNE to automatically design deep neural network architectures. They accomplished this by using a probabilistic grammar with GP and searching for the most efficient network. Other automatic neural network architectures exist such as AlexNet \cite{alexnet}. However they found that ADAG-DNE uses about 2\% of the parameters AlexNet requires, allowing changes to be made more quickly. They therefore created a way to optimise the structure of neural networks for specific tasks by deriving DNN structures encoded in parse trees according to the grammar, collecting samples from the parse trees of the set of good networks which were found after they are decoded and evaluated.

%Activation functions importance within a neural network
\subsection{Activation functions}
Rectified Linear Unit (\textit{ReLU}) is commonly used \cite{hahnloser, jarrett, Nair, xu, hara, dahl, wang} as an activation function in networks with many layers. This is because it overcomes the vanishing gradient problem which many other activation functions such as $sigmoid$, $tan$ and $tanh$ suffer from. However, papers that focus on task-specific activation functions and backing away from using activation functions homogeneously across multiple datasets have found that greater performance can be found. Bingham et al. \cite{bingham} discovered that you can design activation functions on a task-specific basis using a novel evolutionary tree structure design to create customary activation functions. The ability to mutate and swap specific nodes makes this model very versatile and ensures variability. When testing on the CIFAR-10 and CIFAR-100 datasets \cite{krizhevsky}, they found that an activation function developed for one task would indeed transfer to another task, but higher performance can be achieved when a specialised function is discovered specifically for the second task.

Sharma et al’s \cite{sharma} study discusses the importance of activation functions in depth. Their study involved testing off-the-shelf activation functions such as \textit{Sigmoid}, \textit{Tanh}, \textit{Elu}, \textit{ReLU} and \textit{Leaky ReLU} for example against each other. They concluded with several claims. Firstly, for classification problems, a combination of sigmoid functions gives better results. However, due to the vanishing gradient problem, \textit{Sigmoid} and \textit{Tanh} should be avoided. They found that \textit{ReLU} is the most widely used function and performs better than other activation functions in most cases – unless there are dead neurons in the network, in which case \textit{Leaky ReLU} outperforms its counterpart. Sharma et al. also added that \textit{ReLU} must be used in the hidden layer only and not in the output layer. Lastly, and most importantly, they concluded that choosing the right activation function for any particular task or dataset may be a tedious process and may require substantial research as there is no rule of thumb for selecting an activation function because the choice of activation function is dependent on the context/task. 
% Everything has been copied from the previous paper apart from snippets.

\section{Methodology}
As per the standard GE process, a grammar (shown in Figure \ref{fig:grammar}) and the GE phenotype mapping process is used to generate the candidate activation functions. The fittest of these (as determined by how well they work for a neural network applied to the specific classification task) are evolved in a population across generations.

TensorFlow/Keras \cite{keras} was used to create and run the neural networks using TensorFlow's dense layers. A custom activation function is used in the input, hidden and output layers. A tensor and the genome are passed to this activation function. If it is the first epoch and the phenotype has not been created yet, it runs through a recursive function to determine the phenotype. This mapping function is similar to the GE mapping process described by O'Neill et al. \cite{oneill2} which generates the phenotype from the genotype via the grammar.

Essentially, the codons in the genotype determine which of the alternative productions on the right-hand-side grammar rule currently being expanded are chosen to be further expanded in the recursive process.

This process also uses standard wrapping if the codons in the genotype have been exhausted, where the codons being used wrap around to be read from the beginning of the genotype.

Once the string has been generated, a separate function ensures the string is well-formed according to brackets. Then this string is converted into an equation using Python’s built-in `eval’ method \cite{eval}.

A further sanity check involves checking the phenotype for zero division errors. If any are caught, the tensor returned is equal to the original tensor multiplied by 0.0. This acts as the termination of an unfit individual. Lastly, if the phenotype does not contain the original passed tensor then the output of this equation is multiplied by the original tensor element-wise.

Once the activation function has been created, it is passed back to TensorFlow’s sequential model for compiling and the resulting neural network is trained on the training data. After this process has been completed, the fitness of each network in the population is evaluated using the validation accuracy multiplied by the testing F1-score of the predictions. Once the population has been filled, the standard evolutionary process using genetic operators is then continued to evolve the population of networks.

\subsection{Genetic operators}
\indent Mutation is used in evolutionary algorithms to `stir the pot', or to provide some escape from monotony within the population. As previously discussed in section \ref{sec:ge}, Grammatical Evolution is closer to the portrayal of actual DNA than other EA's, and one benefit of this is the silent mutation. In order to make sure these silent mutations exist in this implementation, a random codon is chosen between 0 and the length of the genotype --- even if it is an unused codon. This codon is swapped with another value between 0 and 100. The individual is then mapped to the phenotype, evaluated and tested again before it is placed back into the population.

\indent Crossover is the genetic recombination of genetic information between parents and offspring. Tournament selection is used to determine who will be the parents. Four individuals are chosen out of the population and split into two groups. The individuals from each group are tested, and whichever individual from each group has a higher F-Measure score are chosen to be the parents. Single-point crossover was used because the genotypes are not of variable length. A random point is then chosen between 0 and the length of the genotype. The codons that are between 0 and this point are cloned into a child, whereas the codons that are between the crossover point and the final codon are placed into another child. The same is repeated for both parents, resulting in the recombination of genetic information and the creation of offspring. 

When the offspring are created, they have the potential to fail. This is due to an activation function causing an exploding gradient in which the gradient gets exponentially bigger and causes the loss value of the network to be meaningless, or a vanishing gradient, which results in the loss value of the network producing NaN values. If the creation of the offspring does result in one of these errors, it is simply terminated and the parent of the offspring is placed back into the population. If there is no error, both children are added to the population, and the parents are removed.

\indent Elitism helps keep track of the fittest individual through each generation. If an individual is fitter than the rest of the population and is fitter than the current `elite' individual, it is classed as `elite' and is exempt from mutation and kept in the population until a fitter individual is found.

This method of optimisation can be run once for a large number of generations when a new dataset is being used, to find the optimum activation function. Once the activation function has been found, it can be reused for as long required for the specific task.

\subsection{The Grammar}
The grammar, shown in Figure \ref{fig:grammar}, was designed to allow the production of emergent activation functions. However, the grammar was also designed to allow the possibility of well known activation functions being created, the most well-known of which being \textit{ReLU}. This was used as an example due to the assumption that as ReLU is so popular, it must be the best activation function in at least some scenarios. Experiments will test this assumption later in the paper.

The grammar used was designed to satisfy two conditions: firstly, the number of non-terminal symbols was kept minimal as the genotype was of a linear representation; secondly, we decided to have a minimal amount of symbol biases. Symbols are elements that are either terminal or non-terminal and eventually get mapped to the phenotype. A symbol bias is a bias towards a specific symbol, typically done with prior knowledge of the problem in hand, and often includes a duplication of the symbol in favour. Through testing, we found that too many biases within the grammar led to a lack of variance within the population with respect to the individual's phenotypes. However, we did incorporate a bias to the grammar in general. That is, we designed the grammar to make it difficult for linear activation functions to be designed, as this would create a `lame' individual and waste processing time.

The evolutionary process we designed starts by creating random integer genomes (lists) of size 30 and featuring genes within the range of 0-100. 100 of these are created to fill the population. These genomes are passed one by one into a neural network creation function using the parameters seen in Table~\ref{tab:architecture}.

\begin{figure*}[htb]
\centering
\begin{minipage}{\textwidth}
\begin{lstlisting}[language=json, firstnumber=1, label={fig:grammar}, captionpos=b, caption={The grammar used in Neuvo GEAF to create a single custom activation function. $tensor$ is a terminal node that represents the input tensor entering the activation function.}, mathescape=true]
 <activation_function> ::==  <acti_expr>
 <acti_expr>           ::==  <acti_pre_op>
                        |    <acti_pre_op> <op> <acti_expr>
                        |    "(" <acti_pre_op> <op> <acti_expr> ")"
 <acti_pre_op>         ::==  "tf.math.sin (" <acti_input> ")"
                        |    "tf.math.cos (" <acti_input> ")"
                        |    "tf.math.tan (" <acti_input> ")"
                        |    "tf.math.minimum (" <acti_input> <acti_var> ")"
                        |    "tf.math.maximum (" <acti_input> <acti_var> ")"
                        |    "tf.math.exp (" <acti_input> ")"
                        |    "tf.math.tanh (" <acti_input> ")"
                        |    "tf.math.pow (" <acti_pre_op> <acti_var> ")"
 <acti_input>          ::==  tensor
 <op>                  ::==  + | $\div$ | $\times$ | -
 <acti_var>            ::==  0.1 | 1.0 | 2.0 | 3.0
\end{lstlisting}
\end{minipage}
\end{figure*}

\section{Experimental Setup} \label{exp_setup}
This section discusses the experimental setup. The parameters that define the neural network architecture are shown in Table~\ref{tab:architecture}. All other parameters regarding the neural network such as the learning rate are set to Tensorflow's Sequential model's default parameters.  Evolutionary parameters for this project are found in Table~\ref{tab:evolutionary_par} and the datasets used and their source can be found in Table~\ref{table:3}.
The libraries and hardware specifications used to create Neuvo GEAF are also discussed in this section.

\begin{table}[ht]
\centering
\hskip1.0cm\begin{tabular}{|p{4.0cm}|r|} \hline
\textbf{Parameter description} & \textbf{Parameter} \\ [0.5ex] \hline \hline
Hidden layers & $1-3$ \\ [0.5ex]\hline
Nodes per hidden layer & $8$ \\ [0.5ex]\hline
Optimiser & Adam \\[0.5ex]\hline
Maximum number of epochs & $50$ \\[0.5ex]\hline
Kernel initialiser & Glorot/Xavier uniform \cite{glorot} \\[0.5ex]\hline
Batch size & $4$ \\[0.5ex]\hline
Output layer activation function & Sigmoid\\[0.5ex]\hline
\end{tabular}
\newline
\caption{Neural network creation parameters used throughout this project. Only Sigmoid was used in the output layer to avoid the vanishing gradient problem. The optimal number of hidden layers for each dataset was determined using Neuvo NAS+~\cite{neuvo}. The results were as follows: 1 hidden layer for the Heart disease detection dataset, 1 hidden layer for the Wisconsin Breast Cancer Detection dataset, 3 hidden layers for the Pima Diabetes Detection dataset, and 3 hidden layers for the Sonar Detection dataset.}
\label{tab:architecture}
\end{table}

This model was implemented in Python and Tensorflow on a 64-bit Windows 10 computer using an Intel Core i7-10700k 2.90GHz CPU and an Nvidia GeForce RTX 2070 Super GPU. Tensorflow/Keras~\cite{keras} is an open-source neural network library in Python. TensorFlow uses an object called a tensor that represents a generalised version of a vector that is passed throughout a neural network. 

\begin{table}[ht]
\centering
\hskip1.0cm\begin{tabular}{|p{2.8cm}|r|} \hline
\textbf{Evolutionary para.} & \textbf{Parameter} \\ [0.5ex] \hline \hline
Population size & $100$ \\ [0.5ex]\hline
Generations & $500$ \\ [0.5ex]\hline
Crossover rate & $90\%$ \\[0.5ex]\hline
Mutation rate & $20\%$ \\[0.5ex]\hline
Tournament size & $4$ \\[0.5ex]\hline
Elitism size & $1$\\[0.5ex]\hline
\end{tabular}
\newline
\caption{Evolutionary parameters used for this project.}
\label{tab:evolutionary_par}
\end{table}

Kernel initializer values are set to the Glorot uniform initializer \cite{glorot}, also known as the Xavier uniform initializer. This is to initialize the weights such that the variance of the activations are the same across each layer, helping to reduce the risk of the exploding or vanishing gradient. 

This project also made use of TensorFlow's callbacks function. In particular, to alleviate the problem of fully training a network with a poor phenotype, a callback to stop training early if no significant improvement in the network's loss after 5 epochs was used and this dramatically reduced training speed overall. 

\subsection{Datasets}\label{datasets}
This section describes the datasets that were used in testing and training Neuvo GEAF. Only binary classification datasets were used in this project. However, these varied in size and complexity. Each dataset was shuffled and split 75/25 into training and testing data respectively. This data was then used for 10 evolutionary runs of 500 generations. The output of the neural networks was checked to see if the network output $x < 0.50$ --- if so, the value would be automatically set to 0; if $x \geq 0.50$ it would be set to 1.

For the dataset Sonar~\cite{sonar}, the label values were changed from `M' and `R' to 1 and 0 respectively. Likewise for the Wisconsin breast cancer detection dataset, values were changed from `M' and `B' which represented `Malignant' and `Benign' to 1 and 0.
\begin{table}[h!]
\centering
\resizebox{\columnwidth}{!}{%
\begin{tabular}{|l|r|r|l|} \hline
 \textbf{Name} & \textbf{Attrib.} & \textbf{Instances} & \textbf{Classification} \\ [0.5ex] \hline \hline
 Heart~\cite{heart} & 14 & 303 & Heart disease prediction\\ [0.5ex] \hline
 Pima~\cite{pima} & 9 & 768 & Diabetes detection\\ [0.5ex] \hline
 Sonar~\cite{sonar} & 60 & 208 & Metal or rock\\ [0.5ex] \hline
 WBCD~\cite{breastcancer} & 32 & 569 & Tumour detection\\ [0.5ex] \hline
\end{tabular}%
}
\newline
\caption{Datasets used to evaluate Neuvo GEAF, ordered in ascending order by size (attributes $\times$ instances).}
\label{table:3}
\end{table}
\section{Experimental Results and Discussion}
This section discusses the experimental results. The first experiment's purpose was to see how well a standard artificial neural network on binary classification datasets would perform. The first experiment involved an ANN model using the same features as seen in Table \ref{tab:architecture}. However, instead of evolved activation functions, the model used the popular \textit{ReLU} activation function for the input and hidden layers.

The second experiment involved the creation of task-specific evolvable activation functions using Grammatical Evolution. Both experiments used the same evolutionary parameters as shown in Table~\ref{tab:evolutionary_par} and were tested on the same four binary classification datasets (seen in Table~\ref{table:3}).

\begin{table}[ht]
\centering
\resizebox{\columnwidth}{!}{%
\begin{tabular}{|l|>{\raggedleft\arraybackslash}p{1.2cm}|r|r|r|} \hline
\textbf{Dataset} & \textbf{Training accuracy} & \textbf{MAE} & \textbf{RMSE} & \textbf{F1 Score} \\ [0.5ex] \hline \hline
Heart & 0.688 & 0.258 & 0.508 & 0.800\\ [0.5ex] \hline 
Pima & 0.722 & 0.272 & 0.522 & 0.656\\  [0.5ex] \hline
Sonar & 0.789 & 0.095 & 0.309 & 0.933\\ [0.5ex]\hline
WBCD & 0.873 & 0.034 & 0.186 & 0.941\\ [0.5ex]\hline
\end{tabular}
}
\newline
\caption{The results from individuals that had the best testing F1 Score over 10 individual runs, using the baseline parameters found in Table \ref{tab:architecture}, yet using \textit{ReLU} for the activation functions for the input and hidden layers.}
\label{tab:5}
\end{table}

\begin{table*}[ht]
\resizebox{1.06\textwidth}{!}{
\hskip-1.25cm\begin{tabular}{|r|c|c|c|c|c|} 
 \hline 
 \textbf{Dataset} & \textbf{AF1} & \textbf{AF2} & \textbf{AF3} & \textbf{AF4} & \textbf{AF5}\\ \hline
 WBCD & $tan(x)+cos(x)-$ & $max(x, 2.0)$ & $min(x,0.1)\times$ & $--$& $--$\\ 
 & $tanh(x)-cos(x)$ & & $max(x,0.1)-tanh(x)$ & & \\[1.0ex]  \hline
 Pima & $\dfrac{max(x,0.1)}{sin(x)}$ & $min(x,0.1)$ & $tan(x)$ & $max(x,0.1)$ & $tanh(x)-min(x,0.1)$\\[1.0ex] \hline
 Sonar & $min(x,0.1)$ & $max(x,0.1)-sin(x)$ & $\dfrac{cos(x)}{max(x,2.0)+e^x}$ & $max(x,1.0)+min(x,0.1)$ & $max(x,0.1)-sin(x)$\\[1.0ex] \hline
 Heart & $\dfrac{tanh(x)}{tanh(x)^3}$ & $max(x,1.0)+\dfrac{sin(x)}{max(x,0.1)}$ & $min(x,2.0)-tanh(x)$ & $--$ & $--$\\[1.0ex] \hline
\end{tabular}
}
\newline
\caption{Neuvo GEAF activation functions created after 200 generations on various datasets. These results highlight the fittest individual after each run. The number of activation functions are dependant on the number of hidden layers within each individuals architecture, hence the variation in number of activation functions.}
\label{table:geaf_results}
\end{table*}

Table~\ref{tab:5} displays the binary classification results for the tested datasets, while Table~\ref{tab:comparative_results} compares these results with those obtained from competing machine learning algorithms. Analysing the binary classification results in Table~\ref{tab:5} and Table~\ref{tab:comparative_results} shows that Neuvo GEAF produced activation functions that outperformed ReLU on all datasets. When comparing the results from both tables, improvements in F1-score on testing datasets of between 2.4\% and 9.4\% were found. The results show that catering the activation function for the network's layers will increase the metric which has been chosen for the evolutionary fitness function, in our case F1-score on testing data.

\begin{figure}[h!]
    \centering
    \includegraphics[scale = 0.53]{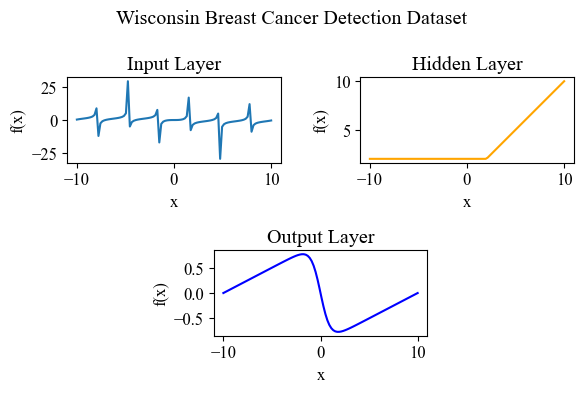}
    \caption{Activation functions produced from the phenotype for the WBCD dataset. Activation functions can be found in Table~\ref{table:geaf_results}. Results are found in Table~\ref{tab:comparative_results}.}
    \label{fig:wbcd_functions}
\end{figure}

\begin{figure}[h!]
    \centering
    \includegraphics[scale = 0.44]{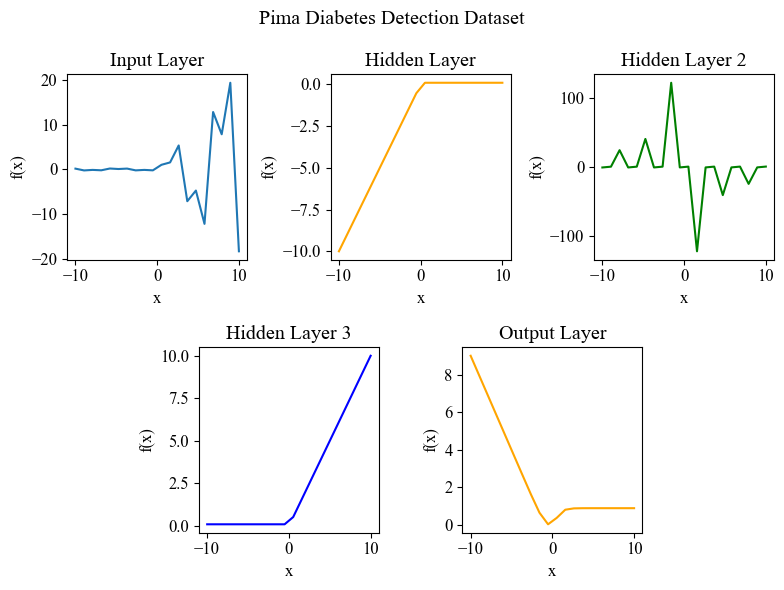}
    \caption{Activation functions produced from the phenotype for the Pima Indian Diabetes Detection dataset. Activation functions can be found in Table~\ref{table:geaf_results}. Results are found in Table~\ref{tab:comparative_results}.}
    \label{fig:pima_functions}
\end{figure}

\begin{figure}[h!]
    \centering
    \includegraphics[scale = 0.56]{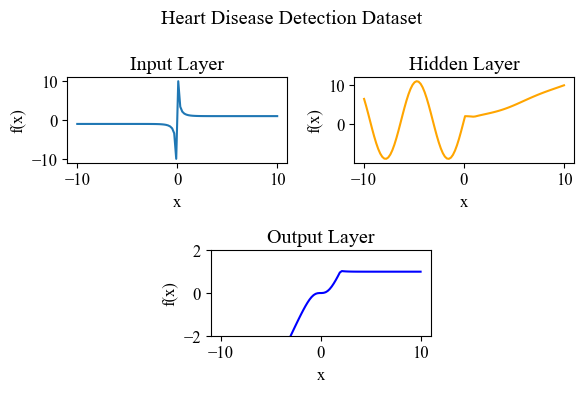}
    \caption{The activation functions generated from the phenotype for the Heart Disease Detection dataset are listed in Table~\ref{table:geaf_results}, with the corresponding results provided in Table~\ref{tab:comparative_results}.}
    \label{fig:heart_functions}
\end{figure}

To clarify, the $x$ axis in the function images found in Figures~[\ref{fig:wbcd_functions}, \ref{fig:heart_functions}, \ref{fig:pima_functions}] are 1000 evenly spaced numbers between -10 and 10. In reality, $x$ would be an input tensor at a neuron level and if one has correctly avoided the vanishing or exploding gradient problems, should typically but not always, contain values that are centered around zero.

As previously discussed, the measure of an individual's fitness was the metric of testing F1-score, thus the fittest individual is the activation functions that produces the highest F1-score. This metric as well as others are shown in Table~\ref{tab:comparative_results}.

The best phenotype for the WBCD dataset consisted of three activation functions; 1.) $tan(x) + cos(x) - tanh(x) - cos(x)$, 2.) $max(x,2.0)$, 3.) $min(x, 0.1) \times max(x, 0.1) -- tanh(x)$. This input layer's activation function here is interesting. It is inherently unstable being between the range of -3 and 3 but it can be quite effective as it does provide non-linearity. The hidden layer's activation function works in tandem with the inherently unstable input layer activation function as it thresholds the output to a non-negative output. The output layer's activation function is extremely interesting as it has created almost a reverse Sigmoid activation function, bounding outputs between -1 and 1, when values of $x$ is in the range [$-20$, $20$].

For the Pima Diabetes Detection dataset, the fittest output layer activation function is very interesting. Coupled with a ReLU type activation function as the previous hidden layer, the output layer has bound itself to the values of $0$ and $x$. This is a very encouraging result and shows a complexity that was produced through emergent properties in the evolution -- that is that each gene has evolved in a symbiosis with previous genes. An example is the activation functions from the Pima Diabetes Detection dataset. The output layer activation function could not properly classify if the previous layers activation function did not bound outputs to values $\geq 0$. The third hidden layer's activation function would suffer deeply from exploding gradients if the previous layers activation function did not bound outputs using a function such as the tangent of $x$ and so on.

The Heart Diseases Detection dataset also produced an interesting phenotype. The hidden layer's activation function can capture complex patterns in the data using the oscillation provided by the $sin(x)$ function whilst also providing some linearity for large values of $x$. This hidden layer is susceptible to exploding gradients. However, the network's output layer activation function smooths the extremely large values using the hyperbolic tangent of $x$ producing a more stable output which is less noisy
. 

One of the great benefits of Grammatical Evolution in our method is that GE offered the ability to map multiple genotypes to the same phenotype, resulting in a greater chance to find an optimal function. Therefore, one can increase the mutation rate and coupled with the idea of silent mutations, increase diversity amongst the population without the risk of consistently producing worse phenotypes after the mutation. 

A further benefit of GE is that complex phenotypes such as the one presented for the WBCD dataset can be created as well as simple functions such as $tan(x)$ present in the Pima datasets due to the recursive nature of our method.

Results show that in 3/4 datasets, Neuvo GEAF has constructed complex output layer activation functions that has the potential to bound values between -1 and 1 when complemented with the outputs from specialised hidden layer activation functions, which is exceptionally promising and shows how effective an algorithm such as Neuvo GEAF can be.

\begin{table}[h!]
\centering
\begin{tabular}{|l|l|r|r|r|r|}  \hline
 \textbf{Algorithm} & \textbf{Metrics} & \textbf{Pima} & \textbf{Heart}
 & \textbf{Sonar} & \textbf{WBCD}\\ [0.5ex]  \hline\hline
 \multirow{4}{*}{GNB} & MAE & 0.108 & 0.066 & 0.141 & 0.023\\ [0.5ex]\cline{2-6}
   & RMSE & 0.232 & 0.182 & 0.265 & 0.107\\ [0.5ex]\cline{2-6}
   & F1 & 0.336 & 0.442 & 0.351 & 0.468\\ \hline 
 \multirow{4}{*}{C4.5} & MAE & 0.129 & 0.089 & 0.107 & 0.019\\ [0.5ex]\cline{2-6}
   & RMSE & 0.254 & 0.211 & 0.231 & 0.098\\ [0.5ex]\cline{2-6}
   & F1 & 0.316 & 0.418 & 0.394 & 0.473\\ [0.5ex]\hline
 \multirow{4}{*}{SVM} & MAE & 0.115 & 0.169 & 0.102 & 0.035\\ [0.5ex]\cline{2-6}
   & RMSE & 0.239 & 0.291 & 0.226 & 0.133\\ [0.5ex]\cline{2-6}
   & F1 & 0.298 & 0.372 & 0.416 & 0.450\\ [0.5ex]\hline
 \multirow{4}{*}{KNN} & MAE & 0.132 & 0.162 & 0.117 & 0.028\\ [0.5ex]\cline{2-6}
   & RMSE & 0.256 & 0.285 & 0.236 & 0.119\\ [0.5ex]\cline{2-6}
   & F1 & 0.300 & 0.355 & 0.408 & 0.462\\ [0.5ex]\hline
 \multirow{4}{*}{ANN} & MAE & 0.443 & 0.473 & 0.336 & 0.211\\ [0.5ex]\cline{2-6}
   & RMSE & 0.472 & 0.489 & 0.396 & 0.278\\ [0.5ex]\cline{2-6}
   & F1 & 0.494 & 0.527 & 0.654 & 0.908\\ [0.5ex]\hline
 \multirow{4}{*}{Neuvo GEAF} & MAE & 0.272 & 0.258 & 0.095 & 0.034 \\ [0.5ex]\cline{2-6}
   & RMSE & 0.522 & 0.508 & 0.309 & 0.186 \\ [0.5ex]\cline{2-6}
   & F1 & \textbf{0.656} & \textbf{0.800} & \textbf{0.933} & \textbf{0.941}\\ [0.5ex]\hline
\end{tabular}%
\caption{Comparison of Neuvo GEAF+ against various other techniques for classifying a variety of binary classification datasets. The individual who achieved the highest F1 score for each dataset is shown in bold font.}
\label{tab:comparative_results}
\end{table}

These results highlight the variation that can be generated using grammatical evolution in the creation of activation functions.

\label{sec:statement}
\section{Conclusion and Future Work}
This paper outlines a novel method of using grammatical evolution to create custom activation functions for neural networks. Results show that our system outperforms the most popular artificial neural network activation function architecture \cite{sharma} on four binary classification datasets, with regards to testing metrics.

It is clear that network parameters must be designed on a task-specific basis to achieve maximum efficiency and accuracy for neural networks. Neuroevolutionary methods like the one presented in this paper provide an unsupervised way to design part of a network's architecture, which is particularly useful when testing on a new unknown dataset.

The phenotypes that were created for all datasets show extremely promising results and show Neuvo GEAF to be a very innovative solution in creating custom activation functions. The results show how creative Neuvo GEAF can be in creating output layer activation functions that suit the task. In this papers case, the task was binary classification problems. Each dataset provided a novel solution to bound its output value so that the network's output would be suitable for binary classification problems.

We hypothesise that this method would be most efficient when used in conjunction with a Neural Architecture Search system such as Neuvo NAS+~\cite{neuvo} whenever a new dataset is to be tested, in order to find the optimum neural network architecture for each parameter of the network. Therefore, future work will test this hypothesis by combining this model with a NEAT system to optimise every parameter within a neural network. We also aim to make optimisation strategies like evolutionary algorithms less time-consuming by combating the time it takes to train each network by adding hyper-parameters such as the number of epochs and the batch size to the individuals genotype as well as using a multi-objective fitness function (speed$\times$accuracy). We also hypothesise that a combination between novel activation functions created by Neuvo GEAF and standardised activation functions could work well.

\bibliography{references}
\bibliographystyle{IEEEtran}

\end{document}